# Degradation Prediction of Semiconductor Lasers using Conditional Variational Autoencoder

Khouloud Abdelli, Helmut Grießer, *Member, IEEE*, Christian Neumeyr, Robert Hohenleitner, and Stephan Pachnicke, *Senior Member, IEEE*

*Abstract*—Semiconductor lasers have been rapidly evolving to meet the demands of next-generation optical networks. This imposes much more stringent requirements on the laser reliability, which are dominated by degradation mechanisms (e.g., sudden degradation) limiting the semiconductor laser lifetime. Physics-based approaches are often used to characterize the degradation behavior analytically, yet explicit domain knowledge and accurate mathematical models are required. Building such models can be very challenging due to a lack of a full understanding of the complex physical processes inducing the degradation under various operating conditions. To overcome the aforementioned limitations, we propose a new data-driven approach, extracting useful insights from the operational monitored data to predict the degradation trend without requiring any specific knowledge or using any physical model. The proposed approach is based on an unsupervised technique, a conditional variational autoencoder, and validated using vertical-cavity surface-emitting laser (VCSEL) and tunable edge emitting laser reliability data. The experimental results confirm that our model (i) achieves a good degradation prediction and generalization performance by yielding an F1 score of 95.3%, (ii) outperforms several baseline ML based anomaly detection techniques, and (iii) helps to shorten the aging tests by early predicting the failed devices before the end of the test and thereby saving costs.

*Index Terms*—Semiconductor lasers, machine learning, degradation prediction, variational autoencoder, semiconductor laser reliability.

## I. INTRODUCTION

SINCE its first invention in 1962, the semiconductor laser has been rapidly developed to meet the needs of optical networks in terms of high launch power, low linewidth, moderate electrical power consumption, and wavelength stability [1,2]. Additionally, for these applications stringent requirements on semiconductor laser reliability exist.

The reliability of such lasers is governed by degradation mechanisms. Several intrinsic and external factors for example contamination, crystal defects, ambient temperature, etc, can adversely affect its performance [3]. Many of these factors are unpredictable and their interaction can lead to complex degradation mechanisms, which are hard to model [4]. Moreover, the complexity of the laser structures, and the diversity of the factors inducing the degradation render it difficult to fully understand the physical processes governing degradation [3]. Therefore, adopting physics-based models that use explicit mathematical equations to model the degradation behavior of the system is challenging.

Recently, data-driven prognostic approaches have been gaining popularity [5-7]. Such methods do not require system specific knowledge or physical models to perform the prognostics and heath monitoring. They learn the system behavior by leveraging the collected data, and thus predict the future state of degradation and the remaining useful life. Developing such prognostic methods requires the availability of run-to-failure data sets, reflecting both, the behavior for normal operation and the degradation process under different operating conditions. However, such data is often unavailable as a device failure is extremely rare during normal system operation, and it would take prohibitively long to generate meaningful reliability data. In order to speed up the degradation process and consequently data collection, often accelerated aging tests are performed by stressing the device with higher operating temperatures or drive currents. The reliability data derived from such accelerated aging tests can be highly unbalanced as the number of degraded or failed device samples is much lower than the number of normal devices. Hence, supervised ML techniques, such as classification-based methods, cannot be used as they require a balanced amount of normal and degraded data. Therefore, unsupervised ML methods, particularly reconstruction-based anomaly detection approaches, are frequently used [8-10]. In this respect, we recently proposed a convolutional autoencoder model for

This work has been performed in the framework of the CELTIC-NEXT project AI-NET-PROTECT (Project ID C2019/3-4), and it is partly funded by the German Federal Ministry of Education and Research (FKZ16KIS1279K).

K. Abdelli is with ADVA Optical Networking SE, Fraunhoferstr. 9a, 82152 Munich/Martinsried, Germany, and with Kiel University (CAU), Chair of Communications, Kaiserstr. 2, 24143 Kiel, Germany (e-mail: kabdelli@adva.com).

H. Grießer is with ADVA Optical Networking SE, Fraunhoferstr. 9a, 82152 Munich/Martinsried, Germany (e-mail: HGriesser@adva.com).

C. Neumeyr and R. Hohenleitner are with Vertilas GmbH, Daimlerstraße 11d, 85748 Munich, Germany (e-mail: neumeyr@vertilas.com, robert.hohenleitner@t-online.de).

S. Pachnicke is with Kiel University (CAU), Chair of Communications, Kaiserstr. 2, 24143 Kiel, Germany (e-mail: stephan.pachnicke@tf.uni-kiel.de).

tunable laser anomaly detection [11]. The ML model was trained with synthetic data generated using generative adversarial networks. Although the ML model can detect anomalies including degradation, it is unable to identify the type of detected anomaly.

In this paper, we go one step further and present an improved framework based on a conditional variational autoencoder for not only predicting the degradation of a semiconductor laser but also for classifying the type of degradation or failure modes (e.g., sudden, gradual degradation). Furthermore, the proposed approach (probabilistic model) provides a slight improvement of 5% compared to our recently presented model [11] (deterministic approach) due to stochastics. The probabilistic modeling of the latent representation by considering its variability helps to capture better and smoother a representation of the input data and thus results in better generalization capability. Furthermore, treating the latent variables as stochastic variables makes the model more flexible and able to handle less probable predictions more effectively and thereby enhances its prediction capability. Our approach is applied to vertical-cavity surface-emitting laser (VCSEL) experimental reliability data derived from accelerated aging tests conducted under different operating conditions as well as to tunable laser reliability data for validation, whereas in [11], the ML model was validated using only tunable laser data and the operating conditions were not provided as an additional input information during the training phase of the model.

The remainder of this paper is structured as follows: Section 2 gives some background information about variational autoencoders (VAEs) and the gated recurrent unit (GRU) algorithm. Section 3 presents the proposed approach. Section 4 describes the experimental data and the validation of the presented method. Conclusions are drawn in Section 5.

## II. BACKGROUND

In this section, we briefly describe the theoretical concepts of the ML algorithms involved in the implementation of the proposed approach.

### A. Autoencoder

An autoencoder is a type of artificial neural network trained in an unsupervised manner to learn reconstructions that are close to its original input [12]. It is composed of two parts, an encoder, and a decoder. The encoder compresses an input $x$ into a lower-dimensional latent-space representation $z$ through a non-linear transformation, and the decoder then maps the encoded representation back into the estimated vector $\hat{x}$ of the original input vector as follows:

$$z = f(Wx + b), \quad (1)$$
$$\hat{x} = g(W'z + b'), \quad (2)$$

where $f$ and $g$ denote the non-linear activation functions of the encoder and the decoder, respectively. The weight matrix $W$ (resp. $W'$) and the bias vector $b$ (resp. $b'$) are the learnable parameters for the encoder (resp. decoder).

The network parameters $\theta = \{W, b, W', b'\}$ are optimized during the training phase by minimizing the reconstruction error $e_r = \|x - \hat{x}\|$, representing the difference between the output $\hat{x}$ and the input $x$.

Autoencoders can come in handy as non-linear feature extraction and dimensionality reduction techniques, which are used to remove the noise or the redundant/correlated or irrelevant information (leading to model overfitting) from the high dimensional input data, while providing a useful lower dimensional representation (encoding) of the input. Then, the encoded features or the compressed representation of the data can be used to train different ML methods (e.g. artificial neural networks, support vector regression, etc) for solving different learning tasks (e.g. classification, regression, etc).

Another application of autoencoders is anomaly detection. Generally, the autoencoders are trained with data representing only the normal behavior. A well-trained autoencoder should be able to reconstruct the normal instances very well by achieving small errors, while it should fail to reproduce the anomalous observations (unseen during the training phase) by yielding much greater reconstruction errors or losses.

### B. Variational Autoencoder

A variational autoencoder is a specific type of deep generative models, having the same architecture as traditional autoencoders [13]. Unlike the conventional autoencoder that deterministically encodes the input into the latent representation and subsequently outputs the reconstructed input, the VAE constrains the latent-space representation $z$ to be distributed according to a prior distribution $p_\theta(z)$, usually a multivariate unit Gaussian distribution, to force the model to learn the input distribution [13]. After that, $x$ is sampled from $p_\theta(x|z)$, which can be derived from a decoder neural network (generative model) with parameters $\theta$, to decode $z$ into a distribution over the observation $x$. Given that the true posterior $p_\theta(z|x)$ is intractable for a continuous latent space $z$, variational inference techniques are often adopted to find an approximate posterior $q_\phi(z|x)$, which can be modeled as an encoder neural network (recognition network)) parameterized by $\phi$. This posterior is usually assumed to be $\mathcal{N}(\mu_\phi(x), \sigma_\phi^2(x))$, where $\mu_\phi(x)$ and $\sigma_\phi(x)$ parameters are to be optimized by the neural network. As shown in Fig.1, a VAE is composed of an encoder, a latent distribution, and a decoder. The input $x$ passes through the encoder (i.e., $q_\phi(z|x)$) to obtain the parameters of the latent distribution (e.g., the mean $\mu$, the standard deviation $\sigma$). Then, the latent vector $z$ is generated by random sampling from the distribution. The decoder (i.e., $p_\theta(x|z)$) then uses the latent vector to produce the reconstruction of the original input $\hat{x}$.

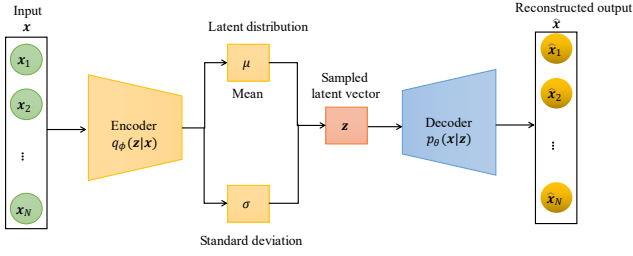

Fig. 1. Structure of the variational autoencoder (VAE).

The encoder and the decoder networks are trained jointly by maximizing the objective function which can be expressed as follows:

$$\mathcal{L}_{VAE}(\theta, \phi; x) = \mathbb{E}_{q_\phi(z|x)}[\log(p_\theta(x|z))] - D_{KL}[q_\phi(z|x) \| p_\theta(z)], \quad (3)$$

where $p_\theta(x|z)$ represents the likelihood of $x$ given $z$, and $\theta$ and $\phi$ are the decoder and the encoder parameters to be optimized.

The first term of eq. (3) is the reconstruction likelihood, i.e, how good the decoder is at learning the training data. The second term of eq. (3) represents the Kullback-Leibler divergence ($D_{KL}$) between the probability distribution learnt by the encoder and the prior distribution of the latent representation $z$, which forces the posterior distribution to be similar to the prior distribution, working as a regularization term.

The aim of training a VAE is to find jointly the optimal $\theta$ and $\phi$ parameters by maximizing the objective function (eq. (3)) with respect to both parameters, whereby maximizing eq. (3) with respect to $\theta$ maximizes the reconstruction likelihood (i.e. reducing the reconstruction error between the input and its reconstruction), whereas the maximization of the objective function with respect to $\phi$ minimizes the $D_{KL}$ (i.e. maximizing the similarity between the true posterior $p_\theta(z|x)$ and its approximation $q_\phi(z|x)$).

A conditional variational autoencoder (CVAE) is an extension of the standard VAE by conditioning the encoder and the decoder with auxiliary covariate information (i.e., additional property of the data such as label or a class) as shown in Fig. 2. The encoder is conditioned on the input $x$ and the covariates $c$ whereas the decoder is conditioned on the latent representation $z$ and $c$. Hence, the objective function of CVAE is expressed as follows:

$$\mathcal{L}_{CVAE}(\theta, \phi; x, c) = \mathbb{E}_{q_\phi(z|x,c)}[\log(p_\theta(x|z,c))] - D_{KL}[q_\phi(z|x,c) \| p_\theta(z|c)], \quad (4)$$

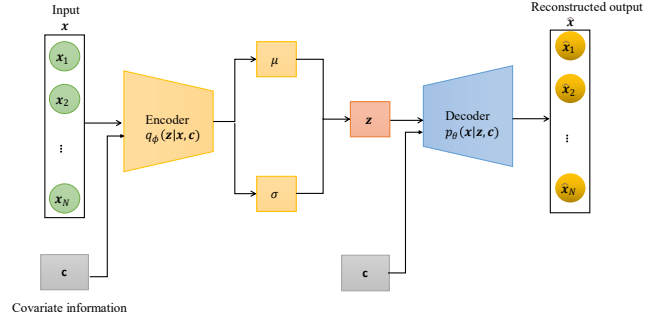

Fig. 2. Structure of the conditional variational autoencoder (CVAE).

*C. Gated Recurrent unit*

The Gated Recurrent Unit (GRU) recently proposed to tackle the gradient vanishing problem [14], is an improved version of standard recurrent neural networks (RNNs), used to process sequential data and to capture long-term dependencies. As shown in Fig. 3, the typical structure of GRU contains two gates, namely reset and update gates, controlling the flow of the information. The update gate regulates the information that flows into the memory, while the reset gate controls the information flowing out the memory.

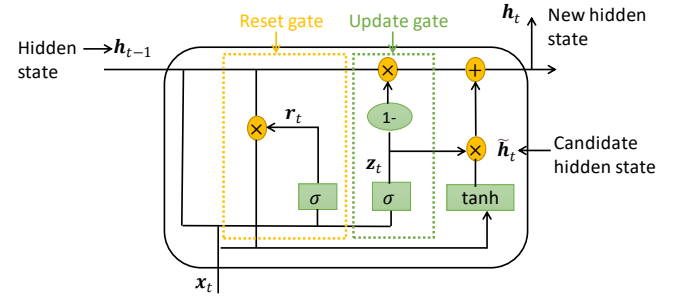

Fig. 3. Structure of the gated recurrent unit (GRU) cell.

The GRU cell is updated at each time step *t* by applying the following equations:

$$z_t = \sigma(W_z x_t + W_z h_{t-1} + b_z), \quad (5)$$
$$r_t = \sigma(W_r x_t + W_r h_{t-1} + b_r), \quad (6)$$
$$\widetilde{h_t} = \tanh(W_h x_t + W_h(r_t \circ h_{t-1}) + b_h), \quad (7)$$
$$h_t = z_t \circ h_{t-1} + (1 - z_t) \circ \widetilde{h_t}, \quad (8)$$

where $z$ denotes the update gate, $r$ represents the reset gate, $x$ is the input vector, $h$ is the output vector and $W$ and $b$ represent the weight matrix and the bias vector, respectively. $\sigma$ is the gate activation function, and tanh represents the output activation function. The "∘" operator represents the Hadamard product.

III. PROPOSED APPROACH

In the following the proposed approach is used for an early identification of degraded or failed devices during the manufacturing phase (i.e., burn-in aging tests) which helps to reduce the time and the costs required to perform those aging tests. Data derived from accelerated or burn-in aging tests has to be available for the training of the ML model.

In addition, the method could be used for early prediction of the degradation or any abnormal behavior that semiconductor lasers deployed in optical networks might exhibit during operation and thus enables early and efficient maintenance plan scheduling. Please note that the ML model in such use case has to be pre-trained with field monitoring data from the normal operating devices, incorporating historical measurements of either the current or the output power values

### A. Model Architecture

The proposed approach is a version of CVAE whereby only the decoder is conditioned on both the latent variable and the operating conditions (auxiliary covariate information), whereas the encoding is independent of the operating conditions. Such approach has a simpler model structure, which requires less computational resources than the standard CVAE. The proposed model is called semi-CVAE (SCVAE) in the following. As shown in Fig. 4, the proposed approach is composed of two sub-models, the encoder, and the decoder. The encoder performs a dimensional compression of the input $x$, incorporating the sequence of output power measurements $[x_0, x_1, ... x_5]$, and outputs the parameters of the latent space distribution, namely the mean vector $\mu$ and the variance vector $\sigma$. The latent vector $z$ is then sampled from the distribution $\mathcal{N}(\mu, \sigma)$. The operating conditions ($oc$) that impact the degradation trend, namely the temperature $T$ and the laser current $I$, combined with $z$ are used to generate a reconstructed input $\hat{x}$ ($[\hat{x}_0, \hat{x}_1, ... \hat{x}_5]$) through the decoder. The objective function of the SCVAE model is formulated as follows:

$$\mathcal{L}_{SCVAE}(\theta, \phi; x, oc) = \mathbb{E}_{q_\phi(z|x)}[\log(p_\theta(x|z, oc))] - D_{KL}[q_\phi(z|x) \parallel p_\theta(z|oc)], \quad (9)$$

The structure of the encoder contains two GRU layers with 40 and 20 cells, respectively, used to extract the features of the sequential input $x$. The decoder is inversely symmetric to the

---

**Algorithm 1: SCVAE based anomaly detection**

**Input**: Normal dataset $(x, oc)$, degraded dataset $(x^{(i)}, oc^{(i)})$ $i = 1, ..., N$, threshold $\alpha$

**Output**: reconstruction error
1:  $\phi, \theta \leftarrow$ train a SCVAE given the normal data $(x, oc)$
2:  **for** $i = 1$ to $N$ **do**
3:      $\mu_z(i), \sigma_z(i) = f_\theta(z | x^{(i)}, oc^{(i)})$
4:      $z^{(i)} \sim \mathcal{N}(\mu_z(i), \sigma_z(i))$
5:      draw one sample $\hat{x}^{(i)}$ from $z^{(i)}$
6:      $e_r(i) = \|x^{(i)} - \hat{x}^{(i)}\|$
7:      **if** $e_r(i) > \alpha$ **then**
8:          $x^{(i)}$ is degraded
9:      **else**
10:         $x^{(i)}$ is normal
11:     **end if**
12: **end for**

---

encoder part. Adaptive moment estimation (Adam) is adopted as an optimizer. The Rectified Linear Unit (ReLU) is selected as an activation function for the hidden layers of the model.

### B. Anomaly Detection

The SCVAE model is trained with only normal data representing the normal operating behavior in order to learn the normal pattern. During the inference phase, the reconstruction error (i.e., square $L_2$ norm) is adopted as anomaly score. Input data with a high anomaly score is regarded as containing anomalies (e.g., sudden degradation, gradual degradation, etc.) because it is expected that the fully trained SCVAE reconstructs normal data with very low reconstruction error, while failing to produce anomalous data it has not been confronted with during the training phase. The process of the classifying an observation as normal or degraded (i.e., anomalous) is summarized in Algorithm 1. If the calculated anomaly score is higher than a threshold α, the instance is classified as "degraded", otherwise it is considered as "normal". α is a parameter to be optimized. Once an anomaly is detected, it is feasible to identify the type of degradation $D_{type}$ (e.g., sudden, or gradual failure modes) by setting a two-threshold system optimized based on the reconstruction errors achieved by the SCVAE for each degradation type as follows:

$$D_{type} = \begin{cases} gradual, & if\ (\ \alpha < e_r \leq \beta) \\ sudden, & if\ (\ e_r > \beta), \end{cases} \quad (10)$$

where $\beta$ is a parameter to be optimized.

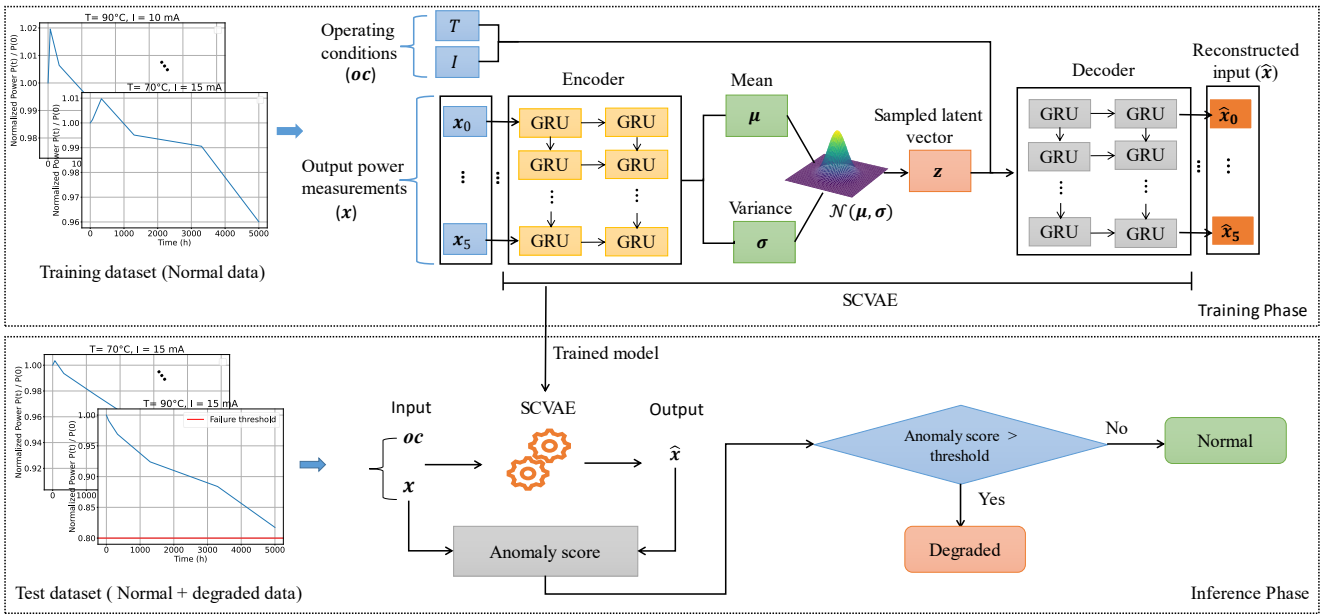

Fig. 4. Proposed GRU-based SCVAE architecture for semiconductor laser degradation prediction.

It is expected that the SCVAE reconstructs the "sudden degradation" samples with higher reconstruction errors compared to the "gradual degradation" samples as the gradual failure mode trend is slow and looks more similar to the normal patterns than the sudden degradation trend that is fast and looks different.

As the patterns of the different degradation types under normal operating conditions might look a bit different from the patterns of the failure models under stressed conditions, the thresholds "α" and "β" for the ML model to be deployed in the field have to be optimized with representative data including the failures of devices observed under normal operating conditions.

## IV. VALIDATION OF THE PROPOSED FRAMEWORK

### A. Experimental Data

To validate the proposed approach, experimental data derived from accelerated aging tests performed for VCSEL devices operating under different operating conditions is used. To strongly increase the laser degradation and thereby speed up the device failure, the aging tests are carried out under high temperature (70°C or 90°C). The output power (i.e., degradation parameter) is monitored for 15,000h under constant current operation. The failure criterion of the device is defined as the decrease of the output power by more than 20% from its initial value. Figure 5 shows examples of aging test results of several semiconductor lasers conducted under different operating conditions. The number of the tested devices shown in Figs. 5 (a), (b), and (c) is 27 (out of which 2 devices failed), 29 (just 2 lasers failed during the aging tests), and 23 (out of which 18 devices broke down) VCSELs, respectively.

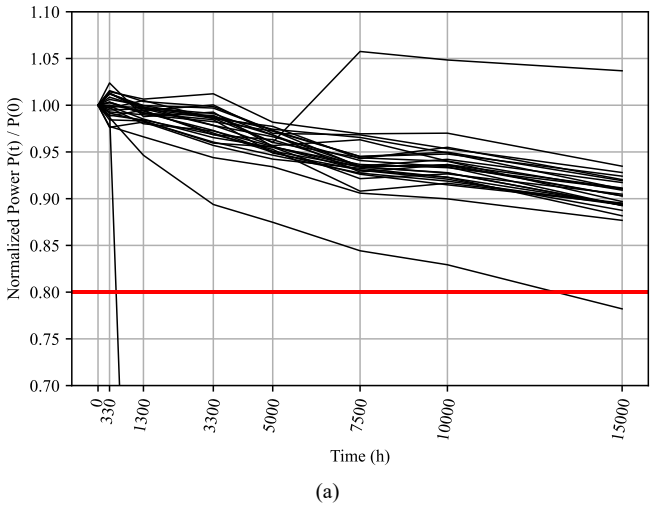

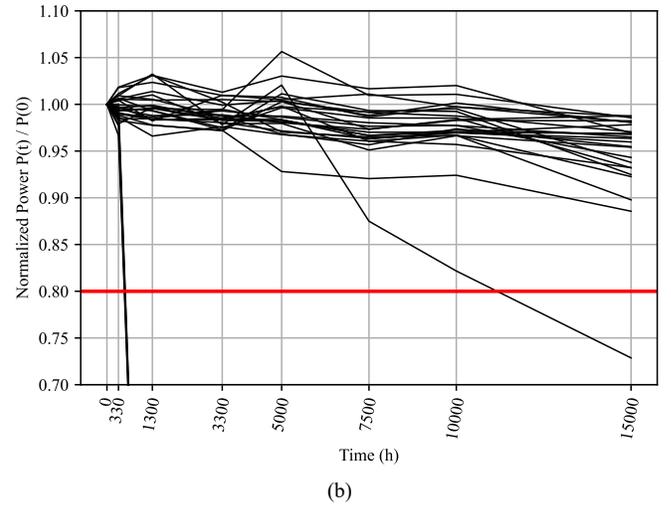

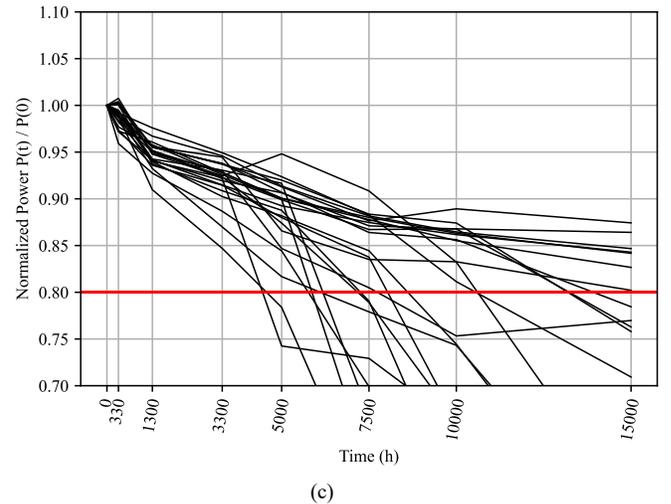

Fig. 5. Experimental aging test data of VCSLs conducted under different operating conditions: (a) $T = 70°C, I = 15$ mA, (b) $T = 90°C, I = 10$ mA, (c) $T = 90°C, I = 15$ mA.

In total, a dataset of 6,786 samples containing the sequences of output power measurements monitored for 5,000 h, combined with the operating conditions $T$ and $I$, is built. The state of the device (normal or degraded) according to the failure criterion defined above is assigned to each sample. The data is then normalized and randomly split into training data (comprising of 80% of the samples) and a test dataset (the remaining 20%). The training dataset includes only samples of normal devices, whereas the test dataset incorporates samples of both, normal (25%) and degraded devices (75%). Only degraded lasers that failed after 5,000 h are considered in order to test the capability of the ML model to predict the degradation early.

### B. Experimental Results

#### 1. Evaluation Metrics

The degradation prediction problem is modeled as a binary classification, whereby we distinguish among sequences with label *"degraded"* (i.e., "positive") or *"normal"* (i.e., "negative"). We consider:

- *true positives (TP)*: Number of "degraded" sequences correctly classified;

- *true negatives (TN)*: Number of "normal" sequences correctly classified;
- *false positives (FP)*: Number of "normal" sequences misclassified as "degraded";
- *false negatives (FN)*: Number of "degraded" sequences misclassified as "normal".

To assess the degradation prediction capability, the following metrics are adopted:

- Precision (P) quantifies the ratio of "degraded" sequences correctly classified and thus the success of the predictions made by the ML model. It is expressed as:

$$P = \frac{TP}{TP + FP}$$

- Recall (R) (also referred as true positive rate) indicates if a high detection rate of "degraded" sequences is achieved at the cost of a high number of sequences wrongly classified as "normal". It is formulated as:

$$R = \frac{TP}{TP + FN}$$

- The F1 score is the harmonic mean of precision and recall and its maximum is the optimum choice for a balance of false positives and false negatives. It is calculated as:

$$F1 = 2\frac{P \cdot R}{P + R}$$

- The false positive rate (FPR) represents the ratio of predicting the "degraded" samples as "normal". It is given by:

$$FPR = \frac{FP}{FP + TP}$$

*2. Performance Evaluation*

The degradation prediction capability of the SCVAE is optimized by selecting a suitable threshold α. Figure 6 illustrates the precision, the recall (i.e. sensitivity), and the F1 score curves as function of α. If the selected threshold is too low, many degraded laser devices will be classified as normal devices, resulting in a higher false positive ratio. Whereas if the chosen threshold is too high, many normal devices will be classified as degraded, leading to a higher false negative ratio. Therefore, the optimal threshold that provides the best precision and recall tradeoff (i.e., maximizing the F1 score) is selected. Please note that if a laser manufacturer might have specific requirements to be considered such as maximizing the recall for an acceptable precision value of 90%, the threshold fulfilling such requirements has to be chosen. For the optimal chosen threshold of 0.013, the precision, the recall, and the F1 score are 98.6%, 92.2%, and 95.3%, respectively.

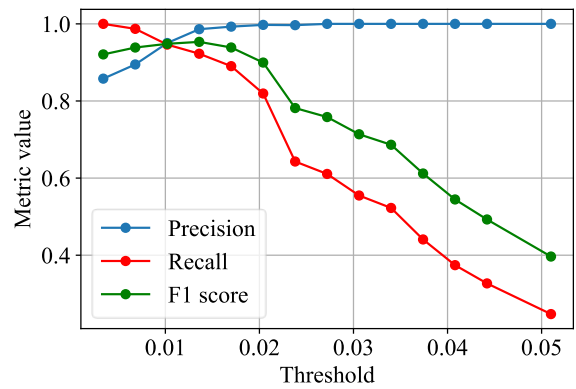

Fig. 6. The optimal threshold selection for anomaly detection based on the precision, recall and F1 scores yielded by SCVAE.

As shown in Fig. 7, the normal and abnormal reconstruction score distributions look quite well separated, which proves that the SCVAE has efficiently learnt the relevant features characterizing the normal behavior or pattern. The receiver operating characteristic (ROC) curve, a probability graph illustrating the performance of the model at various classification threshold settings, which is plotted with the true positive rate on the *y*-axis against the false positive rate on the *x*-axis at different classification thresholds, indicates that the SCVAE model can distinguish very well between the normal and degraded classes by achieving an area under the curve (AUC) (i.e. degree of separability between classes) of 0.96, and that the optimally selected threshold generalizes well for the test dataset.

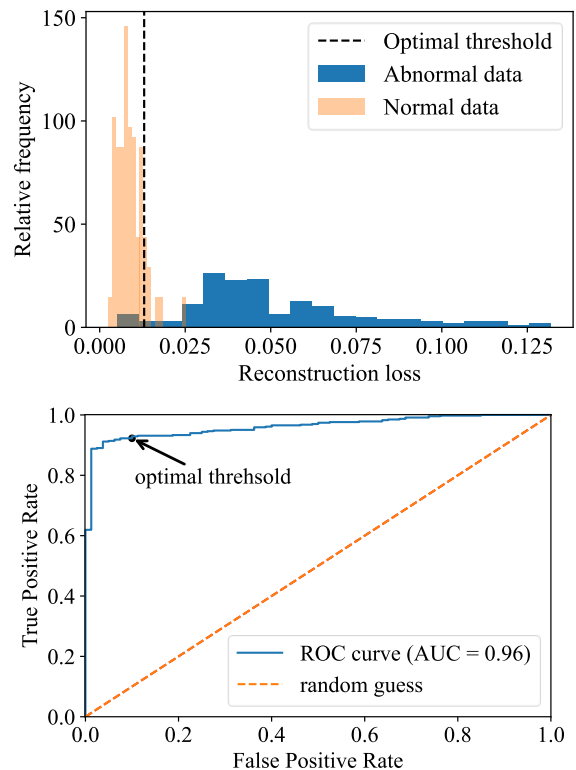

Fig. 7. Representation of the anomaly score distributions (top) and the corresponding receiver operating characteristic (ROC) curve achieved by the SCVAE (bottom).

The feature learning capability of the proposed approach is visually investigated using the t-distributed stochastic neighbor embedding (t-SNE) technique [15]. Figure 8 shows that the learnt features are well separated, which confirms that the proposed model can efficiently distinguish between normal and degraded classes. Please note that as the operating conditions' information is given as an additional input to the decoder for the reconstruction of the input, the SCVAE learns to reconstruct the normal samples for specific operating condition settings as a separate class or cluster. Therefore, for the normal class in Fig. 8 a), two separate clusters are shown there. Fig. 8 (b) illustrates that the latent variables for the normal and degraded classes are of good degree of separability which confirms that the SCAVE can discriminate very well between the two classes.

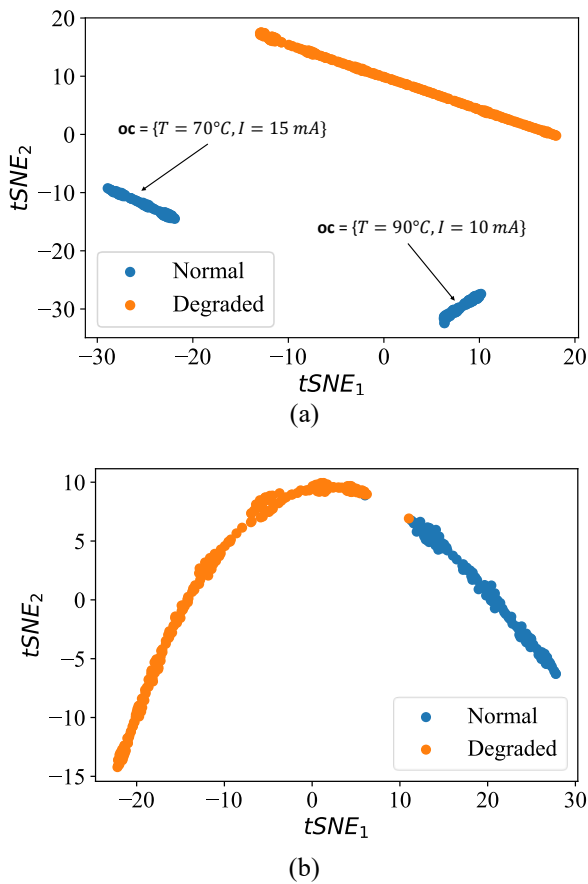

Fig. 8. Visualization of: a) the learned features by SCVAE, and (b) latent variable representations using the t-SNE method.

To evaluate the early degradation prediction capability of the SCVAE model, we consider an unseen test dataset composed of measurement sequences with 5,000 h run time of randomly selected normal and degraded VCSELs that failed after 5,000 h. As shown in Fig. 9, our approach can accurately and early predict the degraded or failed devices before reaching the failure criterion or the end of the aging test (i.e., 15,000 h), which proves the usefulness of the SCVAE in early predicting the failed devices and thus reducing the time and the costs of conducting the aging tests for a long time (the aging time can be shortened by 66.6% while achieving high detection accuracy). Please note that the conventional approaches based on a threshold system for laser failure detection fail to predict the degraded devices at 5,000 h as the decrease of the output power at that time for the considered devices does not reach the failure criterion, whereas the ML model is able to early predict the failed devices. Hence, it is advantageous to adopt an ML model in early predicting degradation.

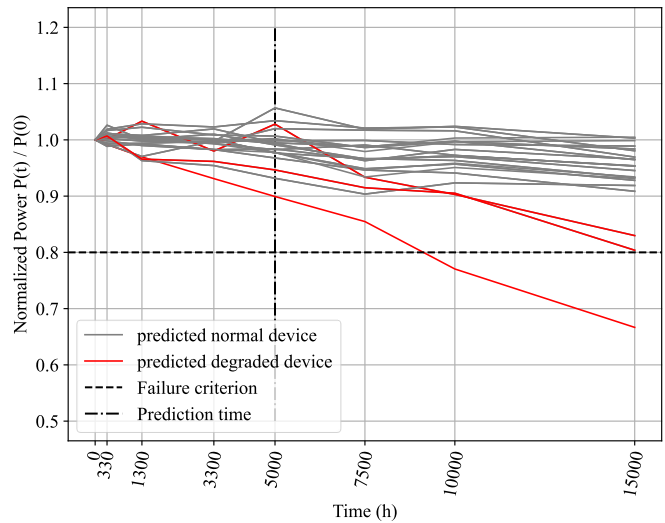

Fig. 9. Assessment of early degradation prediction capability of the SCVAE model.

*2. Optimization of SCVAE*

Adding the information about the operating conditions ($oc$) during the training phase of the SCVAE model has an influence on the performance. The proposed approach including $oc$ during the training stage (i.e., SCVAE with $oc$) is compared with a SCVAE model trained only with sequences of output power measurements $x$ (i.e., SCVAE without $oc$). For the sake of comparison, both models adopt the same architecture (i.e., same encoder and decoder structures). As shown in Table I, including the operating conditions leads to an enhanced degradation prediction capability by achieving an improvement in F1 score of 1.8%. This is expected as $oc$ help the SCVAE model to learn the normal state patterns with different trends under different operating conditions.

TABLE I
IMPACT OF OPERATING CONDITIONS ($oc$) ON THE PERFORMANCE OF SCVAE.

| Method | F1 score (%) |
|---|---|
| SCVAE without $oc$ | 93.5 |
| SCVAE with $oc$ | 95.3 |

In the following the performance impact of training a SCVAE model with output power sequences of different lengths is explored. We trained the SCVAE model with sequences whose lengths varied from 4 to 8.

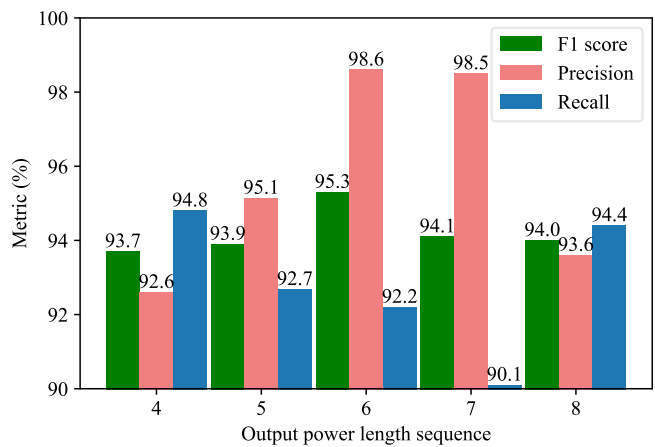

Fig. 10. Impact of output power sequence on the performance of SCVAE.

As depicted in Fig.10, the F1 score and the precision metrics show an ascending trend with the increase of the length of the output power sequence, before reaching the optimum length of 6. Increasing the length helps the SCVAE model to learn more efficiently more relevant features underlying the normal behavior pattern.

*3. Comparison of SCVAE with other ML models*

The proposed SCVAE model is compared with the following anomaly detection-based ML algorithms: One-Class Support Vector machine (OCSVM) [16], Local Outlier Factor (LOF) [17], a standard GRU-based autoencoder (GRU-AE), a convolutional based autoencoder (Conv-AE) [11]. The hyperparameters of the different aforementioned ML models are optimized. OCSVM is with a linear kernel with a coefficient of 0.1 and a parameter denoting an upper bound on the fraction of training errors and a lower bound of the fraction of support vectors, of 0.6. The tuned hyperparameters of LOF are the neighborhood size (defining the neighborhood for the computation of local density) of 100 and the contamination parameter (specifying the proportion of points to be labeled as anomalies) of 0.5. The standard AE is composed of an encoder and a decoder sub-model with 4 layers, whereby the encoder contains two GRU layers with 40 and 20 cells, respectively. And the structure of the decoder is inversely symmetric to the encoder's architecture. The architecture of the convolutional autoencoder is stated in [11], whereby the input of the sequence length is adjusted to 6. The results shown in Fig. 11 demonstrate that our model outperforms the other ML approaches by yielding the highest values of F1 and AUC scores. In comparison, our method achieves significant improvements of more than 5% and 3.3% in AUC and F1 scores, respectively. Compared to the considered baseline (conv-AE, the recently presented model for laser anomaly detection [11]), our approach yields better prediction capability by providing performance improvements of 7% and 4.98% in AUC and F1 score metrics, respectively. The comparison results of the deterministic autoencoders GRU-AE and Conv-AE show that GRU-AE achieves better performance than Conv-AE in terms of AUC and F1 score mainly due to the fact that GRU is better at processing sequential data and to capture relevant features than the convolutional layers.

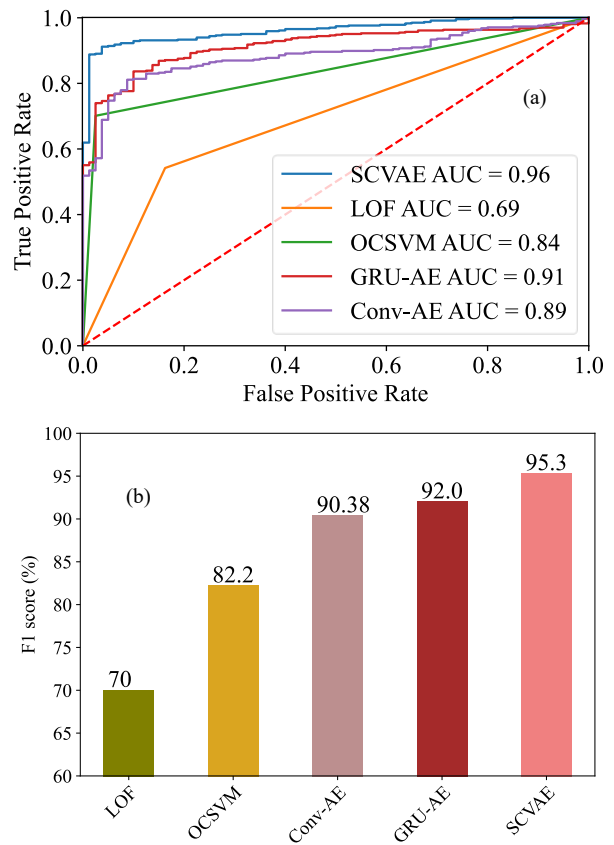

Fig. 11. Comparison of different ML methods in terms of: (a) area under curve (AUC) score, and (b) F1 score.

*4. Robustness Investigation of SCVAE*

To evaluate the robustness of the proposed approach, two unseen test datasets, derived from accelerated aging tests performed on two different batches (wafers) of 16 VCSEL devices conducted under various operating conditions that are either similar or different from the ones of the training dataset, are considered. Let batch 0 denote the batches of the devices used to generate the dataset adopted for testing the proposed model, and batch 1 and batch 2 represent the batches of the lasers utilized to produce the unseen test datasets 1 and 2, respectively. Table II lists the operating conditions of the aging tests performed for each batch.

TABLE II
ROBUSTNESS EVALUATION OF SCVAE MODEL USING UNSEEN DATA DERIVED FROM DIFFERENT BATCHES.

| Batch ID | Operating conditions | F1 score (%) |
|---|---|---|
| 0 | $T$ = 70 °C, 90 °C<br>$I$ = 10 mA; 15 mA | 95.3 |
| 1 | $T$ = 50 °C, 70 °C, 90 °C<br>$I$ = 10 mA; 15 mA | 93.01 |
| 2 | $T$ = 50 °C, 70 °C, 90 °C<br>$I$ = 10 mA, 15 mA; 20 mA | 89.28 |

The results shown in Table II indicate that the SCVAE model still achieves good performance (F1 score higher than 89%) for the unseen test datasets derived from different batches, which proves that the ML model is robust and could generalize well even for the unseen cases with different operating conditions and for different wafers.

*5. Degradation Type Classification*

The capability of SCVAE in classifying the type of the degradation is optimized by selecting an optimal $\beta$. Figure 12 shows the precision, recall and F1 score curves as function of $\beta$. If $\beta$ is too high, many sudden degradation samples will be classified as gradual degradation leading to a low recall value. Whereas if $\beta$ is too low, many gradual degradation failures will be classified as sudden degradation resulting in a low precision value. Therefore, the threshold $\beta$ that maximizes the F1 score is selected as the optimal threshold. For the optimally chosen threshold of 0.1, the precision, the recall, and the F1 score are 96.9%, 94.2%, and 95.5%, respectively.

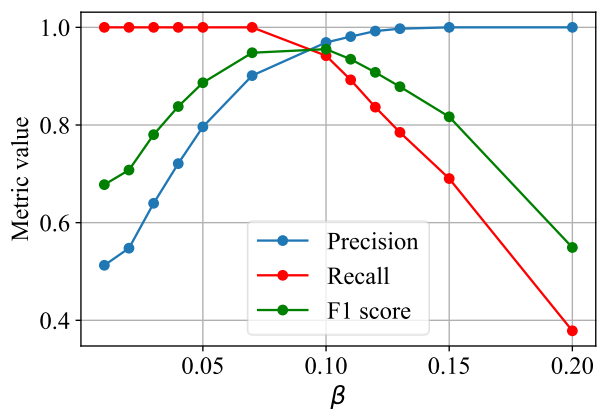

Fig. 12. The optimal threshold selection for degradation type classification based on the precision, recall and F1 scores yielded by SCVAE.

Figure 13 shows that the type of the degradation (i.e., sudden, or gradual) can be identified according to the anomaly score (i.e., the reconstruction error) achieved by the SCVAE. Based on an optimized threshold $\beta$ of 0.1, the anomalous samples (i.e., having anomaly score higher than $\alpha$) can be classified as "gradual" or "sudden" with respect to their anomaly score.

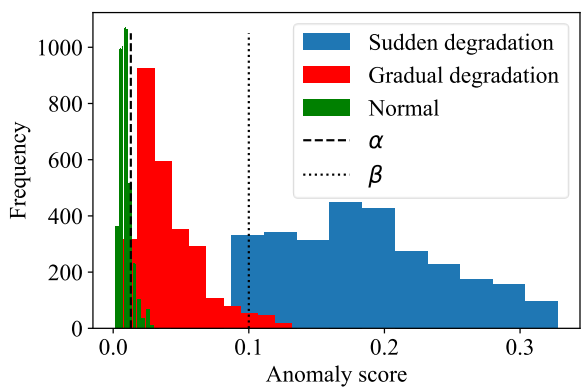

Fig. 13. Degradation type classification according to the anomaly score.

*6. Applicability to other Types of Lasers*

The proposed approach is applied to a different type of lasers (tunable edge emitting lasers). The SCVAE is trained with synthetic data composed of sequences of output power measurements of length 6, and is tested with experimental data. The details about the synthetic data generation and the experimental data can be found in [11]. The SCVAE achieves better performance (F1 score of 94.1%) by providing an improvement of 4.73% in F1 score compared to conv-AE (F1 score of 88.83%).

V. CONCLUSION

We proposed an unsupervised deep learning approach based on a conditional variational autoencoder for semiconductor laser degradation prediction. The proposed model has been tested using VCSEL and tunable edge emitting laser reliability data including normal and degraded devices. The results demonstrated that the proposed method achieves good performance (F1 score of 95.3%) and robustness, and also outperforms other ML algorithms. The approach can help to reduce the required time for aging tests by predicting the failed devices early and thereby saving time and costs.

REFERENCES

[1]  T. Katsuyama, "Development of semiconductor laser for optical communication," SEI Technical Review. 13-20, 2009.
[2]  F. Ujager, Farhan et al., "A review of semiconductor lasers for optical communications," 7th International Symposium on High-Capacity Optical Networks and Enabling Technologies, HONET 2010.
[3]  J. Jimenez, Juan, "Laser diode reliability: Crystal defects and degradation modes," Comptes Rendus Physique, 2003.
[4]  P. Koomsap, "Condition monitoring and lifetime estimation of a CO2 laser," Journal of Laser Applications - J LASER APPL. 15, 2003.
[5]  Z. Liu, et al, "Similarity-Based Difference Analysis Approach for Remaining Useful Life Prediction of GaAs-Based Semiconductor Lasers," IEEE Access, vol. 5, pp. 21508-21523, 2017.
[6]  K. Abdelli, et al., "Lifetime Prediction of 1550 nm DFB Laser using Machine learning Techniques," OFC (2020).
[7]  K. Abdelli et al., "A Hybrid CNN-LSTM Approach for Laser Remaining Useful Life Prediction," OECC, 2021.
[8]  C. Zhou, Chong et al., "Anomaly Detection with Robust Deep Autoencoders." Proceedings of the 23rd ACM SIGKDD International Conference on Knowledge Discovery and Data Mining (2017).
[9]  T. Chen, et al., "Unsupervised Anomaly Detection of Industrial Robots Using Sliding-Window Convolutional Variational Autoencoder," in IEEE Access, vol. 8, pp. 47072-47081, 2020.
[10] T. Luo and S. G. Nagarajan, "Distributed Anomaly Detection Using Autoencoder Neural Networks in WSN for IoT," 2018 IEEE International Conference on Communications (ICC), 2018, pp. 1-6.
[11] K. Abdelli, et al., "A Machine Learning-based Framework for Predictive Maintenance of Semiconductor Laser for Optical Communication", IEEE J. Lightw. Technol. (JLT), 2022.
[12] M.A. Kramer, "Nonlinear principal component analysis using autoassociative neural networks," AIChE Journal. 37 (2): 233–243, 1991.
[13] D.P. Kingma, et al., "Auto-encoding variational Bayes," ICLR (2014).
[14] K. Cho et al, "Learning Phrase Representations using RNN Encoder-Decoder for Statistical Machine Translation," Proceedings of the 2014 Conference on Empirical Methods in Natural Language Processing, 1724-1734, 2014.
[15] Van der Maaten et al., ''Visualizing data using t-SNE,'' Journal of Machine Learning Research. 9. 2579-2605, 2008.
[16] B. Scholkopf et al., "Estimating the support of a high-dimentional distribution," Neural Computation 13 (2001) 1443–1471.
[17] M. M Breunig et al., "LOF: identifying density-based local outliers," In ACM sigmod record, Vol. 29. ACM, 93—104, 2000.